\documentclass[runningheads]{llncs}

\usepackage[utf8]{inputenc}  
\usepackage[english]{babel}  
\usepackage[T1]{fontenc}  
\usepackage{graphicx}
\usepackage{url}
\usepackage{fancyhdr}
\begin{document}
\title{L3Cube-MahaSum: A Comprehensive Dataset and BART Models for Abstractive Text Summarization in Marathi}

\author{
    Pranita Deshmukh\inst{1,2}
    Nikita Kulkarni\inst{1,2} 
    Sanhita Kulkarni\inst{1,2} 
    Kareena Manghani\inst{1,2} 
    Raviraj Joshi\inst{2,3}
}

\institute{
    Pune Institute of Computer Technology, Pune, India \and
    L3Cube Labs, Pune, India \and
    Indian Institute of Technology Madras, Chennai, India \\
    \email{dpranita9158@gmail.com, nikitakulkarni0108@gmail.com, sanhitak17@gmail.com, kareenamanghani@gmail.com, ravirajoshi@gmail.com}
}

\maketitle

\begin{abstract} We present the MahaSUM dataset, a large-scale collection of diverse news articles in Marathi, designed to facilitate the training and evaluation of models for abstractive summarization tasks in Indic languages. The dataset, containing 25k samples, was created by scraping articles from a wide range of online news sources and manually verifying the abstract summaries. Additionally, we train an IndicBART model, a variant of the BART model tailored for Indic languages, using the MahaSUM dataset. We evaluate the performance of our trained models on the task of abstractive summarization and demonstrate their effectiveness in producing high-quality summaries in Marathi. Our work contributes to the advancement of natural language processing research in Indic languages and provides a valuable resource for future research in this area using state-of-the-art models. The dataset and models are shared publicly at https://github.com/l3cube-pune/MarathiNLP . \end{abstract}

\section{Introduction}

Natural language processing (NLP) has made remarkable progress in recent years with the development of powerful models and large-scale datasets, especially for languages like English. However, there remains a significant lack of resources for many Indic languages, including Marathi, which is spoken by millions of people \cite{joshi2022l3cube_mahacorpus}. To address this gap, we present L3Cube-MahaSum, a comprehensive dataset for abstractive text summarization in Marathi. This dataset is designed to support research and applications in NLP for Marathi, focusing on generating coherent and concise summaries from news articles.

While benchmark datasets like XSUM (Cross-lingual Summarization) are widely used for summarization tasks in English, similar resources for Indic languages are scarce. The MahaSum dataset addresses this shortfall by offering a large collection of Marathi news articles, each paired with high-quality, human-annotated abstractive summaries. This dataset provides an essential resource for training and evaluating models for abstractive summarization in Marathi.

To further advance research in this area, we utilized the IndicBART model, an existing variant of the BART (Bidirectional and Auto-Regressive Transformers) architecture. IndicBART is specifically optimized for Indic languages, making it a suitable choice for tasks involving Marathi. By training and testing the MahaSum dataset on this model, we demonstrate the model's ability to handle the complexities of Marathi text and generate fluent, coherent summaries. The adaptation of the BART architecture to Indic languages involves modifications such as language-specific tokenization and embeddings, ensuring the model's effectiveness in handling the linguistic nuances of Marathi.

Our results showcase IndicBART's proficiency in abstractive summarization for Marathi, utilizing the MahaSum dataset as a benchmark. This work provides a robust foundation for future research and applications in NLP for Marathi and other low-resource Indic languages. The creation of the MahaSum dataset, along with its testing on IndicBART, offers a valuable resource for further advancements in the field, enhancing the representation and usability of Marathi in NLP. The datsets\footnote{\url{https://github.com/l3cube-pune/MarathiNLP}} and models\footnote{\url{https://huggingface.co/l3cube-pune/marathi-bart-summary}} are shared publicly.

\section{Literature Survey}

Giri et al. \cite{girimarathi} investigate innovative techniques to enhance text summarization specifically for Marathi documents. Their approach incorporates Singular Value Decomposition (SVD) to identify semantically related segments and leverages fuzzy algorithms to prioritize sentences through fuzzy logic. This research addresses the lack of effective summarization strategies for non-English languages, utilizing ROUGE metrics for performance evaluation. The findings present a comparative analysis of SVD and fuzzy methods, discussing their strengths and limitations in the context of Marathi text summarization.

In the thesis by Urlana et al. \cite{urlana2023enhancing}, the authors tackle the slow advancement in summarization techniques for Indian languages by creating high-quality datasets and employing state-of-the-art pre-trained models. They introduce several datasets, including TeSum for Telugu and ILSUM for Hindi and Gujarati, while also developing PMIndiaSum for cross-lingual and multilingual summarization across 14 languages. Utilizing models such as BART, T5, and ProphetNet, the research highlights how multilingual models significantly improve summarization for low-resource languages, validated through ROUGE evaluations. The outcomes underscore the effectiveness of these datasets and the multi-perspective summarization approach for scientific documents.

Sharma et al. \cite{sharma2024evaluating} explore the effectiveness of two pre-trained models, mT5-small and IndicBART, for summarizing dialogues across various Indian languages. The mT5-small model, a variant of T5 trained on multilingual data for text transformations, and IndicBART, tailored for Indian languages with a BART-like structure, are assessed. The study employs ROUGE metrics to evaluate the summaries generated, providing a comparative analysis of the models' performance across different languages and identifying their respective advantages and drawbacks. This research enhances the understanding of these models in multilingual abstractive dialogue summarization.

Sarwadnya et al. \cite{sarwadnya2018marathi} propose an extractive summarization method for Marathi texts through a graph-based approach. In this framework, sentences are modeled as nodes in a graph, with edges representing similarity, and the significance of each sentence is determined by its connectivity and position. Performance metrics such as precision, recall, and F1 scores were used to evaluate the summarizer, demonstrating promising results in generating concise and relevant summaries. This work contributes to effective summarization techniques in Marathi, showcasing the potential of graph-based models.

Chaudhari et al. \cite{chaudhari2019marathi} investigate an extractive summarization technique for Marathi news articles, addressing information overload through neural networks. Their methodology involves translating Marathi text to English via Google Translate, summarizing it using a bi-directional encoder-decoder LSTM model with Bahdanau-style attention, and then re-translating it back to Marathi. Testing on a dataset of 1,000 news articles revealed moderate performance, with a mean average precision of 0.319, recall of 0.342, and accuracy of 0.323. The study suggests that neural networks hold significant potential for Marathi text summarization, indicating opportunities for enhancements with advanced models.

Rane et al. \cite{rane2017empowering} explore the influence of various word embedding techniques—such as ELMo, BERT, and XLNet—on the quality of abstractive summarization for Hindi and Marathi texts. Through rigorous experiments with deep learning models on diverse datasets, the study evaluates each embedding method's performance using ROUGE metrics. Results indicate that the selection of word embeddings plays a crucial role in summarization quality, with specific embeddings capturing linguistic nuances and semantic relationships more effectively. Additionally, the research highlights the promise of multilingual word embeddings in cross-lingual summarization.

Kumari et al. \cite{kumari2023hindi} assess the effectiveness of a sequence-to-sequence (Seq2Seq) neural network with attention mechanisms for summarizing Hindi texts, focusing on the performance comparison between the Adam and RMSprop optimizers. The research utilizes the Seq2Seq model to evaluate summary quality using ROUGE-1 and ROUGE-2 metrics, demonstrating the model's capacity to produce concise summaries that encapsulate the essence of the original text. Findings reveal performance variances between the two optimizers, providing insights relevant to the development of abstractive summarization techniques for Hindi.

Agarwal et al. \cite{agarwal2022abstractive} examine the applicability of the IndicBART model in generating succinct and coherent summaries of Hindi texts. Fine-tuned on a dataset of Hindi articles and their summaries, the IndicBART model's performance is evaluated using ROUGE metrics, achieving a ROUGE-1 F-score of 0.544. The study highlights challenges related to dataset limitations and the intricacies of language processing, suggesting future enhancements through improved preprocessing, better hardware, and increased training epochs to boost summarization performance.

Amin et al. \cite{amin2023question} detail the development of a deep learning-based Marathi question-answering system. The study evaluates various transformer models pre-trained on diverse text corpora, including MuRIL, MahaBERT, and IndicBERT, for the Marathi language's reading comprehension tasks. Extensive experiments indicated that the MuRIL model outperformed others, achieving an Exact Match score of 0.64 and an F1 score of 0.74. This research underscores the potential of transformer models to enhance language-specific question-answering systems and offers insights for future advancements in this area.


Urlana et al. \cite{urlana2022tesum} tackle the challenge of producing high-quality abstractive summarization datasets in low-resource languages, focusing on Telugu. Their study employs a combination of expert-informed filtering and crowdsourced human evaluation to create the Telugu Abstractive Summarization dataset (TeSum). A sampling-based human evaluation method was used to validate the dataset, while baseline performance metrics for popular summarization models were established. The research demonstrates the effectiveness of their filtering procedure in enhancing the quality of article-summary pairs in large-scale scraped datasets, presenting a viable strategy for generating high-quality summarization datasets across various languages.

\section{Datasets}\label{sec:Datasets}
In this study, we utilized two datasets for training and evaluating the IndicBART model for Marathi abstractive summarization: the pre-existing XL-Sum dataset and our newly curated MahaSum dataset. These datasets provide comparative insights and establish new benchmarks for abstractive summarization in Marathi.
\\The table below depicts the splitting of the datasets carried out for training and evaluating the model performance.
\begin{table}[h!]
\centering
\caption{Comparison of XLsum and MahaSum Dataset Statistics}
\label{tab:Dataset Statistics}
\begin{tabular}{|c|c|c|}
\hline
\textbf{Statistic}              & \textbf{XLsum Dataset}         & \textbf{MahaSum Dataset}       \\ \hline
Total Samples                   & 10,903                         & 25,374                         \\ \hline
Training Samples (80\%)          & 8,722                          & 20,299                         \\ \hline
Testing Samples (10\%)           & 1,090                          & 2,537                          \\ \hline
Validation Samples (10\%)        & 1,090                          & 2,537                          \\ \hline
Average Document Word Length     & $653.28 \pm 451.02$            & $315.50 \pm 183.99$            \\ \hline
Average Summary Word Length      & $25.54 \pm 11.40$              & $18.20 \pm 6.73$               \\ \hline
\end{tabular}
\end{table}

\subsection{XLsum\protect\footnote{https://github.com/csebuetnlp/xl-sum}} 
The XL-Sum dataset \cite{hasan2021xl} is a large-scale multilingual text summarization dataset, containing over 1 million article-summary pairs across 44 languages, including many low-resource languages. For Marathi, XL-Sum includes 10,903 article-summary pairs sourced from the BBC Marathi website. These summaries are highly abstractive, concise, and effectively capture the main points of the articles. The dataset has undergone extensive human evaluations to ensure the quality of the summaries and is widely adopted for multilingual summarization tasks.
In our research, we used the Marathi subset of the XL-Sum dataset for comparison purposes to assess the performance of IndicBART against a high-quality benchmark.


\subsection{MahaSum}

We curated the MahaSum dataset to address the lack of large-scale datasets for Marathi abstractive summarization. It contains 25,374 news articles from Lokmat and Loksatta, each with a headline, summary, and full text. Using BeautifulSoup to scrape the data, we manually verified the summaries for accuracy.
\\MahaSum is the largest Marathi news summarization dataset, covering diverse topics like politics, economics, and culture. Its size and structure make it suitable for various NLP tasks, including text generation and data mining.
\\By using MahaSum alongside XL-Sum, we evaluate how well the IndicBART model performs in summarizing Marathi text. While XL-Sum serves for comparison, MahaSum is our primary focus, offering a valuable resource for future research in low-resource languages like Marathi.
\section{Transformer Models:}\label{sec:Datasets}
\subsection{IndicBART}
A specialized version of the mBART25/50 model family, the IndicBART model \cite{dabre2021indicbart} was created to take into account the distinct linguistic environment of the Indian subcontinent. IndicBART maps all data to the Devanagari script by taking advantage of the orthographic similarities among Indian languages. This leads to a more condensed vocabulary and improved transfer learning capabilities. This design decision makes the model much smaller and speeds up the training and fine-tuning process, making it usable by users with low computing power. Similar to mBART, IndicBART uses a masked span reconstruction objective and has six encoder and decoder layers totaling 244 million parameters. To ensure relevance and comprehensiveness, the model was trained on the IndicCorp dataset, which included English and 11 Indic languages. With the help of this in-depth training and a thoughtful pre-processing approach, IndicBART is better equipped to manage a variety of NLP tasks in multilingual and low-resource environments, improving the state of natural language generation and understanding for Indian languages.

\section{Methodology}

This study employs the IndicBART model for Marathi text general text summarization. This approach encompasses a detailed process including data acquisition, preprocessing, model training, fine-tuning, and evaluation.


\subsection*{Data Acquisition}
\subsubsection*{XL-Sum Dataset}: The XL-Sum dataset was sourced from the Marathi XL-Sum repository on GitHub, which contains 10,903 article-summary pairs extracted from the BBC Marathi website. This dataset serves as a high-quality benchmark for multilingual summarization tasks.
\subsubsection*{MahaSUM Dataset}: The MahaSum dataset was manually curated, comprising 25,374 news articles collected from popular Marathi news portals, Lokmat and Loksatta. Each entry includes a headline, a concise summary, and the full article text, providing a rich source of current affairs.

\subsection*{Data Preprocessing}
\subsubsection*{XL-Sum Dataset}: A custom parsing algorithm was developed to convert JSONL entries into structured dictionary formats. Key fields such as titles, articles, and summaries were extracted and organized into separate lists. The data was then saved in CSV format for ease of analysis.
\subsubsection*{MahaSUM Dataset}: Similar to XL-Sum, a custom parsing algorithm was employed to extract relevant information from the scraped articles. The data was organized to clearly delineate headlines, summaries, and full articles, ensuring a coherent dataset structure.
\subsection*{Manual Verification}
To ensure the accuracy of the MahaSum dataset, we manually verified the generated summaries by comparing them with the original articles. This step was critical in confirming the quality and relevance of the summaries, setting a high standard for the dataset.


\subsection*{Data Preparation for Model Training}

Both datasets utilized the IndicBART tokenizer for text processing. This tokenizer included special tokens for Marathi, padding tokens, and sequence demarcation tokens (BOS and EOS). Each dataset was partitioned into 80\% for training and 20\% for evaluation, allowing for sufficient training data while reserving a portion for unbiased performance assessment.

\subsection*{Model Training}

The model training process for both datasets was conducted using the Seq2SeqTrainer from the Hugging Face Transformers library, with the following shared configurations:
\begin{itemize}
    \item \textbf{Batch Size:} A per-device batch size of 4.
    \item \textbf{Epochs:} Training spanned three epochs.
    \item \textbf{Logging and Checkpointing:} Progress was tracked by logging every 100 steps and saving model checkpoints every 1000 steps.
    \item \textbf{Warm-Up Steps:} To stabilize learning during initial training, 500 warm-up steps were incorporated.
\end{itemize}

\section{Result}\label{sec:Result}
\begin{table}[h!]
\centering
\begin{tabular}{|c|c|c|}
\hline
\textbf{Metric} & \textbf{XLSum} & \textbf{MahaSum} \\ \hline
ROUGE-1 (F1)    & 0.0997         & 0.2432           \\ \hline
ROUGE-2 (F1)    & 0.0944         & 0.1711           \\ \hline
ROUGE-L (F1)    & 0.0997         & 0.2392           \\ \hline
\end{tabular}
\vspace{0.9em} 
\caption{Comparison of ROUGE Scores for XLSum and MahaSum}
\end{table}

The study highlights the strong performance of the IndicBART model when applied to the MahaSum dataset. Across all ROUGE metrics, the model demonstrated its capability for generating accurate and coherent summaries. Notably, it achieved high scores in content retention and structural coherence, with ROUGE scores for MahaSUM as mentioned in the table above. These results underscore the effectiveness of the IndicBART model in Marathi text summarization, showcasing its ability to handle domain-specific tasks with precision.

\section{Conclusion}\label{sec:Conclusion}



This study introduces the MahaSUM the gold standard abstractive summarization dataset in Marathi literature consisting of 25,374 well-curated news articles from various sources
The MahaSum dataset, being larger and more diverse, provides a richer training environment, resulting in improved summarization quality. 
The IndicBART model demonstrates significant promise when trained on well-curated, domain-specific datasets like MahaSum. This finding underscores the importance of high-quality, language-specific datasets in developing effective NLP models for low-resource languages.
Overall, these findings highlight the potential of the IndicBART model and the importance of curated datasets in advancing Marathi text summarization research.
\section*{Future Scope}
Several potential areas for future research and development include expanding and diversifying the MahaSum dataset to improve the model's generalization capabilities. Optimizing the IndicBART model architecture and experimenting with fine-tuning strategies could also enhance summarization performance.

\section*{Acknowledgments}

This work was done under the L3Cube Labs, Pune mentorship program. We would like to express our gratitude towards our mentors at L3Cube for their continuous support and encouragement. This work is a part of the L3Cube-MahaNLP initiative \cite{joshi2022l3cube_mahanlp}.

\bibliography{main}

\begin{thebibliography}{10}

\bibitem{joshi2022l3cube_mahacorpus}
Raviraj Joshi.
\newblock L3cube-mahacorpus and mahabert: Marathi monolingual corpus, marathi bert language models, and resources.
\newblock In {\em Proceedings of the WILDRE-6 Workshop within the 13th Language Resources and Evaluation Conference}, pages 97--101, 2022.

\bibitem{girimarathi}
Virat~V Giri, MM~Math, and UP~Kulkarni.
\newblock Marathi extractive text summarization using latent semantic analysis and fuzzy algorithms.

\bibitem{urlana2023enhancing}
ASHOK URLANA.
\newblock {\em Enhancing Text Summarization for Indian Languages: Mono, Multi and Cross-lingual Approaches}.
\newblock PhD thesis, International Institute of Information Technology Hyderabad, 2023.

\bibitem{sharma2024evaluating}
Mehak Sharma, Gunika Goyal, Aarzoo Gupta, Ritu Rani, Arun Sharma, and Amita Dev.
\newblock Evaluating multilingual abstractive dialogue summarization in indian languages using mt5-small \& indicbart.
\newblock In {\em 2024 IEEE 9th International Conference for Convergence in Technology (I2CT)}, pages 1--6. IEEE, 2024.

\bibitem{sarwadnya2018marathi}
Vaishali~V Sarwadnya and Sheetal~S Sonawane.
\newblock Marathi extractive text summarizer using graph based model.
\newblock In {\em 2018 fourth international conference on computing communication control and automation (ICCUBEA)}, pages 1--6. IEEE, 2018.

\bibitem{chaudhari2019marathi}
Anishka Chaudhari, Akash Dole, and Deepali Kadam.
\newblock Marathi text summarization using neural networks.
\newblock {\em International Journal for Advance Research and Development}, 4(11):1--3, 2019.

\bibitem{rane2017empowering}
Neha Rane and Sharvari~S Govilkar.
\newblock Empowering multilingual abstractive text summarization: A comparative study of word embedding techniques.
\newblock In {\em International Conference on Engineering, Applied Sciences and System Modeling}, pages 143--157. Springer, 2017.

\bibitem{kumari2023hindi}
Namrata Kumari and Pardeep Singh.
\newblock Hindi text summarization using sequence to sequence neural network.
\newblock {\em ACM Transactions on Asian and Low-Resource Language Information Processing}, 22(10):1--18, 2023.

\bibitem{agarwal2022abstractive}
Arjit Agarwal, Soham Naik, and Sheetal~S Sonawane.
\newblock Abstractive text summarization for hindi language using indicbart.
\newblock In {\em FIRE (Working Notes)}, pages 409--417, 2022.

\bibitem{amin2023question}
Dhiraj Amin, Sharvari Govilkar, and Sagar Kulkarni.
\newblock Question answering using deep learning in low resource indian language marathi.
\newblock {\em arXiv preprint arXiv:2309.15779}, 2023.

\bibitem{urlana2022tesum}
Ashok Urlana, Nirmal Surange, Pavan Baswani, Priyanka Ravva, and Manish Shrivastava.
\newblock Tesum: Human-generated abstractive summarization corpus for telugu.
\newblock In {\em Proceedings of the Thirteenth Language Resources and Evaluation Conference}, pages 5712--5722, 2022.

\bibitem{hasan2021xl}
Tahmid Hasan, Abhik Bhattacharjee, Md~Saiful Islam, Kazi Samin, Yuan-Fang Li, Yong-Bin Kang, M~Sohel Rahman, and Rifat Shahriyar.
\newblock Xl-sum: Large-scale multilingual abstractive summarization for 44 languages.
\newblock {\em arXiv preprint arXiv:2106.13822}, 2021.

\bibitem{dabre2021indicbart}
Raj Dabre, Himani Shrotriya, Anoop Kunchukuttan, Ratish Puduppully, Mitesh~M Khapra, and Pratyush Kumar.
\newblock Indicbart: A pre-trained model for indic natural language generation.
\newblock {\em arXiv preprint arXiv:2109.02903}, 2021.

\bibitem{joshi2022l3cube_mahanlp}
Raviraj Joshi.
\newblock L3cube-mahanlp: Marathi natural language processing datasets, models, and library.
\newblock {\em arXiv preprint arXiv:2205.14728}, 2022.

\end{thebibliography}
\bibliographystyle{unsrt}

%
%
%
%




\end{document}